\documentclass[10pt,conference]{IEEEtran}
\usepackage{comment}

\usepackage{amsmath}
\usepackage{amssymb}
\usepackage{graphicx}
\usepackage[T1]{fontenc}
\usepackage[utf8]{inputenc}
\usepackage{booktabs}
\usepackage{hyperref}
\usepackage{url}
\usepackage{cite}
\usepackage{microtype}
\usepackage{tabularx}
\usepackage{xcolor}
\usepackage{xspace}
\usepackage{hyperref}
\usepackage[skip=4pt]{caption}
\usepackage{subcaption}

\usepackage{colortbl}
\definecolor{halluccolor}{RGB}{220, 50, 50}
\definecolor{faithcolor}{RGB}{30, 130, 60}

\title{Not All Claims Are Equally Risky: FACTOR for Adaptive Verification in Factual Long-Form Generation}

\author{
\IEEEauthorblockN{
Areeba Hassan$^{*}$,
Arooj Kausar$^{*}$,
Syeda Kisaa Fatima$^{*}$,
Gibrail Islam, 
Mehwish Fatima
}

\IEEEauthorblockA{
School of Electrical Engineering and Computer Science (SEECS),\\
National University of Sciences and Technology (NUST),\\
Islamabad, Pakistan\\
\{ahassan.msai25seecs, akausar.msai25seecs, sfatima.msai25seecs, gibrail.islam, mehwish.fatima\}@seecs.edu.pk
}

\IEEEauthorblockA{
$^{*}$Equal contribution \\
$^{\dagger}$Corresponding author: mehwish.fatima@seecs.edu.pk
}

}
\begin{document}
\maketitle

\begin{abstract}
Large Language Models (LLMs) generate fluent long-form text, however, often add unsupported factual claims. Existing verification techniques improve factuality by grounding generation in external evidence. However, the same verification policy usually applies to all claims despite being differences in hallucination risks. We propose \textit{FACTOR} (\textit{FACTuality-Oriented Risk-aware Verification}), an inference-time model that adapts verification criteria according to claim-level uncertainty. FACTOR combines uncertainty estimation, adaptive language inference verification, and candidate re-ranking to allocate verification effort where it is most needed. We evaluate \textit{FACTOR} on FactScore benchmark showing that adaptive verification improves factuality while reducing verification cost simultaneously. We further perform different ablation studies to identify the primary driver of these gains. Our results show the effective and model-agnostic performance of \textit{FACTOR} for improving factuality in long-form generation.
\end{abstract}

\section{Introduction}
Large language models (LLMs) generate fluent long-form text, yet factual reliability remains a major challenge~\cite{ji2023survey, mallen2023trust, min2023factscore}. This limitation is particularly evident in knowledge-intensive tasks such as biography generation, question answering, and report writing, where responses consist of multiple factual claims that must remain grounded in verifiable evidence~\cite{min2023factscore, lewis2020retrieval}. A biography of Marie Curie, for example, may correctly identify her as a Nobel Prize winner while simultaneously generating an incorrect birth year or a fabricated affiliation. Such hallucinations reduce the trustworthiness of coherent outputs and remain a key problem for deployment of LLMs-based factual applications \cite{ji2023survey}.

Existing approaches improve factuality through retrieval and verification. Retrieval-Augmented Generation (RAG) grounds generation in external evidence rather than relying solely on parametric memory \cite{lewis2020retrieval}. However, retrieved evidence does not guarantee faithful generation, as models may introduce unsupported details or drift away from the retrieved context \cite{shi2023distract,mallen2023trust}. Natural Language Inference (NLI)-based verification further improves factuality by filtering unsupported claims \cite{laban2022summac}, while uncertainty estimation methods use signals such as token entropy and semantic consistency to identify unreliable generations \cite{kadavath2022language,malinin2021uncertainty,manakul2023selfcheckgpt}. Yet existing systems apply the same verification policy to all claims, even though hallucination risk varies substantially within a response.

\begin{table}[t]
\centering
\scriptsize
\setlength{\tabcolsep}{3pt}
\renewcommand{\arraystretch}{1.1}
\begin{tabular}{p{0.95\columnwidth}}
\toprule
\textbf{Source:}
\textit{Marie Curie was a Polish-born physicist and chemist who conducted
pioneering research on radioactivity. Born on 7 November 1867 in Warsaw,
she was the first woman to win a Nobel Prize, receiving the Physics Prize
in 1903 jointly with Pierre Curie and Henri Becquerel, and later received
the Nobel Prize in Chemistry in 1911.}
\\
\midrule
\textbf{Zero-shot:}
\textit{Marie Curie was born in
{\color{halluccolor}\textbf{1876}} in Warsaw. She won the Nobel Prize in
{\color{halluccolor}\textbf{Chemistry in 1904}} and later joined
{\color{halluccolor}\textbf{MIT as a professor of physics}}.
}

\textbf{Incorrect:} birth year (1876 $\neq$ 1867); Nobel field and year
(Chemistry 1904 $\neq$ Physics 1903); MIT affiliation fabricated.
FactScore $\approx$ 7\%.
\\
\midrule
\textbf{FACTOR:}
\textit{Marie Curie was born on
{\color{faithcolor}7 November 1867} in Warsaw. She conducted pioneering
research on radioactivity and became
{\color{faithcolor}the first woman to win a Nobel Prize}, awarded in
{\color{faithcolor}Physics in 1903} jointly with Pierre Curie and Henri
Becquerel. She later received the
{\color{faithcolor}Nobel Prize in Chemistry in 1911}.}

\textbf{Correct:} All claims verified against retrieved passages. High-uncertainty claims
routed to strict two-passage NLI gate.
FactScore $\approx$ 42\%.
\\
\bottomrule
\end{tabular}
\caption{Representative biography generation example showing why verification should be adaptive. The zero-shot output is fluent but contains unsupported claims, whereas FACTOR retains supported facts and filters hallucinated ones.}
\label{tab:intro_example}
\end{table}

This limitation is visible in Table~\ref{tab:intro_example}. The zero-shot biography preserves fluency but corrupts multiple factual details, whereas adaptive verification retains the supported claims and removes unsupported ones. The example shows that not all claims require the same level of scrutiny: high-risk claims such as dates and affiliations need stricter verification, while lower-risk claims can be verified more lightly. We therefore propose \textbf{FACTOR} (\textit{FACTuality-Oriented Risk-aware Verification}), an inference-time framework that adapts verification criteria according to claim-level uncertainty. FACTOR combines retrieval, uncertainty estimation, adaptive NLI verification, and candidate re-ranking to allocate verification effort where it is most needed. We evaluate FACTOR on the FactScore benchmark \cite{min2023factscore} against strong baselines. Our contributions are threefold: (1) we introduce a risk-aware verification framework, (2) we combine token-level entropy with semantic consistency for routing, and (3) we show that adaptive verification improves factuality while reducing verification overhead.

\section{Related Work}
This section reviews prior work on retrieval and verification for factual generation, uncertainty estimation for hallucination detection, and the FactScore benchmark used for evaluation.

\subsection{Retrieval and Verification for Factual Generation}
Retrieval-Augmented Generation (RAG) improves factuality by grounding generation in external evidence rather than relying solely on parametric memory \cite{lewis2020retrieval}. However, retrieved evidence does not guarantee faithful generation, as models may introduce unsupported details or drift away from the retrieved context during generation \cite{shi2023distract,mallen2023trust}. To address this, many approaches use Natural Language Inference (NLI) verification to evaluate if generated claims are supported by retrieved evidence \cite{laban2022summac}. NLI-based verification has proven to be useful in factual consistency assessment and claim verification, with cross-encoder architectures like DeBERTa showing excellent entailment performance \cite{he2021deberta}. Nonetheless, most of the verification systems implement a rigid verification policy to all generated claims.

\subsection{Uncertainty Estimation and Hallucination Detection}
Uncertainty signals give us an alternative viewpoint on factuality. Previously, token-level entropy and distributional uncertainty are shown to correlate with model confidence and generation reliability \cite{malinin2021uncertainty}. Similarly, SelfCheckGPT measures semantic consistency across multiple sampled generations and demonstrates that disagreement between samples is often associated with hallucination \cite{manakul2023selfcheckgpt}. Other studies show that language models possess some degree of uncertainty awareness, although calibration remains imperfect in open-ended generation settings \cite{kadavath2022language}. These methods primarily use uncertainty for hallucination detection rather than verification.

\subsection{Factuality Evaluation}
We evaluate factuality using the FactScore benchmark \cite{min2023factscore}, which measures the proportion of generated atomic facts supported by external evidence. FactScore has become a standard benchmark for long-form factual generation because it evaluates claim-level correctness rather than surface-level text similarity. The benchmark supports both human and NLI-based verification and reveals substantial factuality gaps even for strong proprietary language models.

\begin{figure}[ht]
    \centering
    \includegraphics[width=0.40\textwidth]{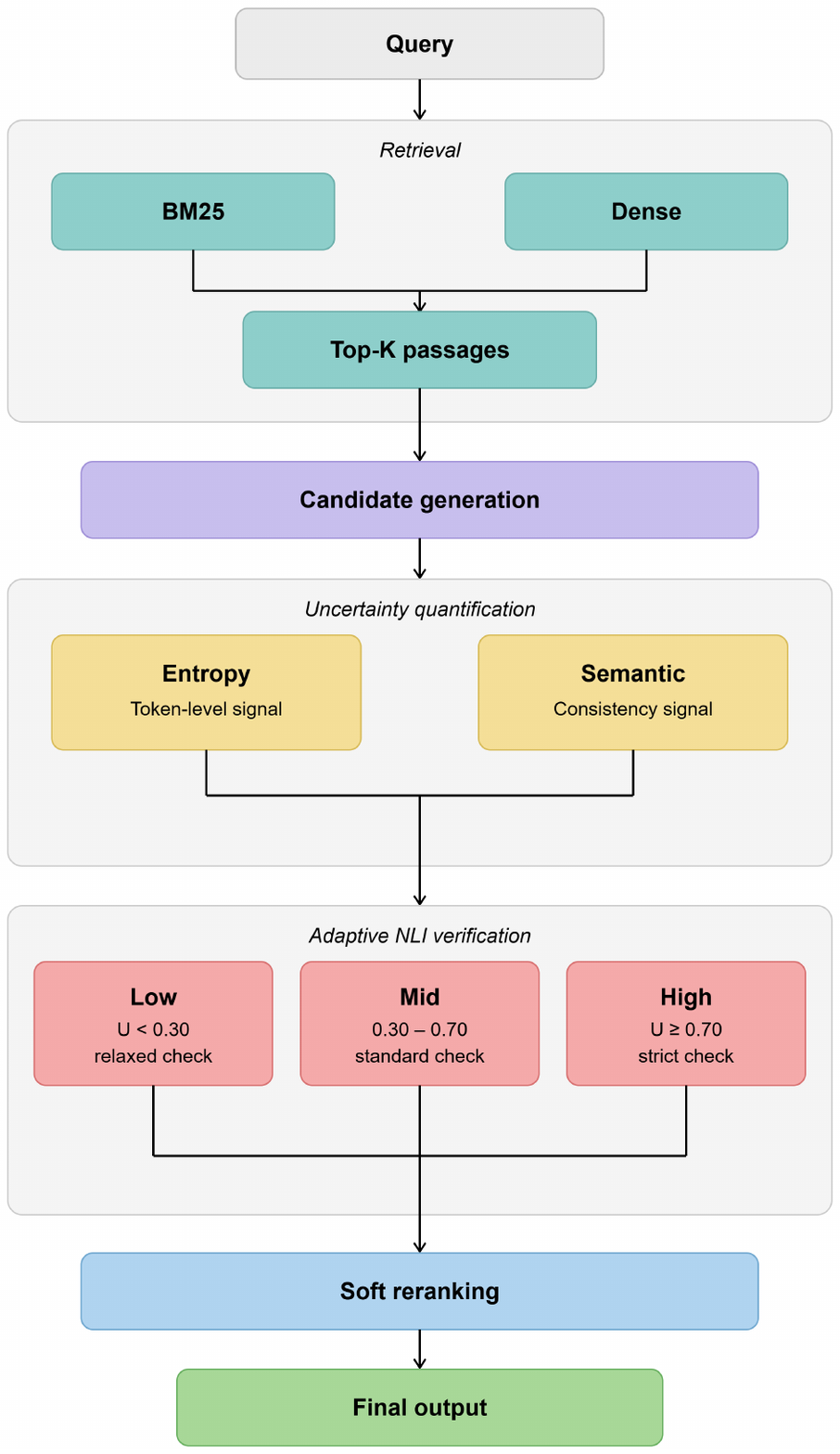}
    \caption{Overview of FACTOR architecture.}
    \label{fig:ua_acd_pipeline}
\end{figure}

\section{FACTOR: FACTuality-Oriented Risk-aware Verification}
Normally, verification-based factuality systems use a uniform verification stringency for all generated claims. However, the claims in a generated text have varying hallucination risk. Some claims are better supported by retrieved evidence, whereas others exhibit higher levels of uncertainty and are more likely to be incorrect. FACTOR addresses this limitation by devising an adaptive verification criteria as per claim-level uncertainty in the response. Figure~\ref{fig:ua_acd_pipeline} presents the overall architecture.

\subsection{Hybrid Retrieval}
FACTOR uses a hybrid retrieval strategy that combines BM25 \cite{robertson2009probabilistic} and dense retrieval to retrieve evidence from Wikipedia. BM25 captures lexical overlap between the query and candidate passages, while dense retrieval computes semantic similarity. Combination of these signals improves evidence coverage and prevents retrieval failures due to paraphrasing or entity ambiguity. The retrieval score is computed as:

{\small
\begin{equation*}
\text{score}(p, q) = \frac{1}{2} \times \text{score}_{\text{BM25}}(p, q) + \frac{1}{2} \times \text{score}_{\text{Dense}}(p, q)
\end{equation*}
 }
where $p$ denotes a candidate passage and $q$ denotes the query. We retain the top-$k$ passages and apply an entity-aware filtering to mitigate risks of cross-entity contamination.

\subsection{Candidate Generation}
FACTOR uses and evidence-first prompt (evidence being the retrived passages) to generate four candidate biographies. All candidates have an identical prompt. the responses differ only because of stochastic sampling. Multiple candidate generation increases the likelihood of producing a response that is diverse as well as faithful to the retrieved evidence.
 
\subsection{Uncertainty-Aware Verification}
FACTOR verifies the generated claims based on their characterizing uncertainty estimate. First, the framework quantifies uncertainty. Then, it adaptively determine the verification stringency for each claim based on the estimated uncertainty.

\subsubsection{Uncertainty Estimation}
Claims have different chances of being incorrect. FACTOR estimates this risk using token-level entropy and semantic consistency, which are complementary. Token-level entropy measures the spread of the model's next-token distribution:

{\small \begin{equation*}
H_t = -\sum_i p_i \log p_i
\end{equation*}}
where $p_i$ is the probability assigned to token $i$. Higher entropy is indicative of higher uncertainty during generation.

Token-level confidence alone does not guarantee factual correctness. We therefore estimate semantic consistency by sampling two additional continuations from the same prompt and measuring agreement across outputs using Sentence-BERT embeddings. Claims that vary substantially across samples receive higher uncertainty scores. We combine both signals into a single uncertainty estimate:

{\small \begin{equation*}
U = \frac{1}{2} \times U_{\text{entropy}} + \frac{1}{2} \times U_{\text{consistency}}
\end{equation*}}
where $U_{\text{entropy}}$ and $U_{\text{consistency}}$ denote normalized entropy and consistency scores, respectively.

\subsubsection{Adaptive Verification}
FACTOR uses uncertainty to determine verification criteria. Low-risk claims undergo lightweight verification, whereas uncertain claims must satisfy stricter evidence requirements. Verification is performed using a DeBERTa-based NLI cross-encoder. Table~\ref{tab:2} summarizes the verification policies for different uncertainty levels.

\begin{table}[h]
\centering
\scriptsize
\begin{tabular}{lcc}
\toprule
\textbf{Tier} & \textbf{$U$ Range} & \textbf{NLI Threshold} \\
\midrule
Low & $U < 0.30$ & 0.60 \\
Mid & $0.30 \leq U < 0.70$ & 0.75 \\
High & $U \geq 0.70$ & 0.85 \\
\bottomrule
\end{tabular}
\caption{Adaptive verification policies used by FACTOR. Verification criteria increases with estimated claim uncertainty.}
\label{tab:2}
\end{table}

For high-uncertainty claims, FACTOR additionally requires support from multiple retrieved passages. This routing strategy concentrates verification effort on claims that are most likely to hallucinate.
 
\subsection{Candidate re-ranking}
FACTOR ranks candidate biographies using both factuality and fluency. We estimate factuality through the average entailment score across all claims:

{\small
\begin{equation*}
\frac{1}{|C|}
\sum_{c \in C}
\max(\text{NLI}_{\text{entail}}(c)),
\end{equation*}}
where $C$ denotes the set of claims in a candidate biography.

We estimate fluency using perplexity:

{\small \begin{equation*}
f = \frac{1}{1 + \text{PPL} / 100}
\end{equation*}}

We combine both signals into a final re-ranking score:

{\small \begin{equation*}
\alpha \bar{e}
+
(1-\alpha)f,
\end{equation*}}
where $\alpha=0.6$. This weighting mechanism gives priority to factuality while also preserving fluency of output and quality. FACTOR selects the candidate with the highest score as the final output.

\section{Experiments}
\subsection{Dataset}
We evaluate FACTOR on a 50-entity subset of the FactScore biography benchmark \cite{min2023factscore}, sampled using a fixed random seed. We use Wikipedia as the evidence corpus and retrieve passages from the March 2022 Wikipedia snapshot. For each entity, we retain up to 20 passages containing approximately 100 words each. All methods access the same retrieval corpus to ensure a fair comparison.

\subsection{Models}
We compare FACTOR against three baselines: Zero-shot Generation, Standard RAG, and Static Verification. All methods use the same generator (Phi-2), embedding model (\texttt{all-MiniLM-L6-v2}), and NLI verifier (\texttt{nli-deberta-v3-small}).\footnote{Models are available at: 
\texttt{Phi-2}: \href{https://huggingface.co/microsoft/phi-2}{huggingface.co/microsoft/phi-2}, 
\texttt{all-MiniLM-L6-v2}: \href{https://huggingface.co/sentence-transformers/all-MiniLM-L6-v2}{huggingface.co/sentence-transformers/all-MiniLM-L6-v2}, and \texttt{nli-deberta-v3-small}: \href{https://huggingface.co/cross-encoder/nli-deberta-v3-small}{huggingface.co/cross-encoder/nli-deberta-v3-small}.} We keep all generation settings fixed across conditions to isolate the effect of adaptive verification. 

Table~\ref{tab:3} summarizes the generation parameters.
\begin{table}[ht]
\centering
\begin{tabular}{lc}
\toprule
\textbf{Parameter} & \textbf{Value} \\
\midrule
Temperature & 0.80 \\
Top-$p$ & 0.92 \\
Repetition penalty & 1.15 \\
Max new tokens & 256 \\
Min new tokens & 80 \\
Candidates & 4 \\
\bottomrule
\end{tabular}
\caption{Generation parameters used across all experiments.}
\label{tab:3}
\end{table}

FACTOR generates four candidate biographies per entity and applies uncertainty-aware verification and re-ranking during inference. All experiments run on a Kaggle T4 GPU with 16 GB VRAM.

\textbf{Zero-shot Generation.} The model generates a biography without retrieval or verification.

\textbf{Standard RAG.} Retrieved passages are prepended to the prompt, but no claim verification is performed.

\textbf{Static Verification.} Retrieved passages are used during generation, and all claims are verified using a fixed NLI threshold. This baseline differs from FACTOR only in the absence of uncertainty-aware adaptation.

\subsection{Evaluation Metrics}
We evaluate factuality using FactScore \cite{min2023factscore}, which measures the proportion of generated claims supported by retrieved evidence. We additionally report hallucination rate, mean NLI entailment score, verification cost, and generation latency.

To assess statistical significance, we compare per-entity FactScore distributions using paired $t$-tests and Wilcoxon signed-rank tests with a significance level of $\alpha=0.05$.

\begin{table}[ht]
\centering
\scriptsize
\setlength{\tabcolsep}{3pt}
\renewcommand{\arraystretch}{1.15}
\resizebox{\columnwidth}{!}{%
\begin{tabular}{@{}lccccc@{}}
\toprule
\textbf{Method} & \textbf{Fact Score $\uparrow$} & \textbf{Hall.\ $\downarrow$} & \textbf{NLI $\uparrow$} & \textbf{Words $\downarrow$} & \textbf{p (Wil.) $\downarrow$} \\
\midrule
Zero-shot Gen.  & $6.5 \pm 9.9$            & 93.5 & 7.5  & 159 & $<0.001$ \\
Standard RAG    & $19.7 \pm 17.6$          & 80.3 & 22.4 & 185 & $<0.001$ \\
Static Constr.  & $36.8 \pm 19.8$          & 63.2 & 38.2 & 173 & $0.040$  \\
FACTOR          & $\mathbf{42.3 \pm 21.9}$ & \textbf{57.7} & \textbf{41.1} & \textbf{163} & $\text{---}^{\dagger}$ \\
\bottomrule
\end{tabular}%
}
\caption{Results on the 50-topic FactScore benchmark. FACTOR achieves the strongest factuality and lowest hallucination rate among all methods. $^{\dagger}$$p$-values (Wilcoxon signed-rank test) reflect pairwise comparisons of each baseline against FACTOR; a self-comparison is undefined for FACTOR, hence ``---''.}
\label{tab:4}
\end{table}

\section{Results}
Table~\ref{tab:4} compares FACTOR against zero-shot generation, standard RAG, and static verification. FACTOR achieves the highest factuality and lowest hallucination rate across all methods. The improvement over Static Verification demonstrates that adapting verification criteria according to claim-level uncertainty is more effective than applying a uniform verification policy.

The gap in performance between Zero-shot Generation and Standard RAG is confirmed by External evidence for factual generation. Retrieval alone does not, however, remove hallucinations, as unsubstantiated claims go on. Can appear even when relevant passages are available. Static Verification further enhances the factuality by filtering out content which is not supported,however this fixed verification policy results in equal treatment of all claims. The Baseline is the scenario without FACTOR, and is consistently outperformed by FACTOR.According to a verification effort that is most effective when allocated according to estimated the risk of hallucination.

\begin{figure}[ht]
    \centering
    \includegraphics[width=\columnwidth]{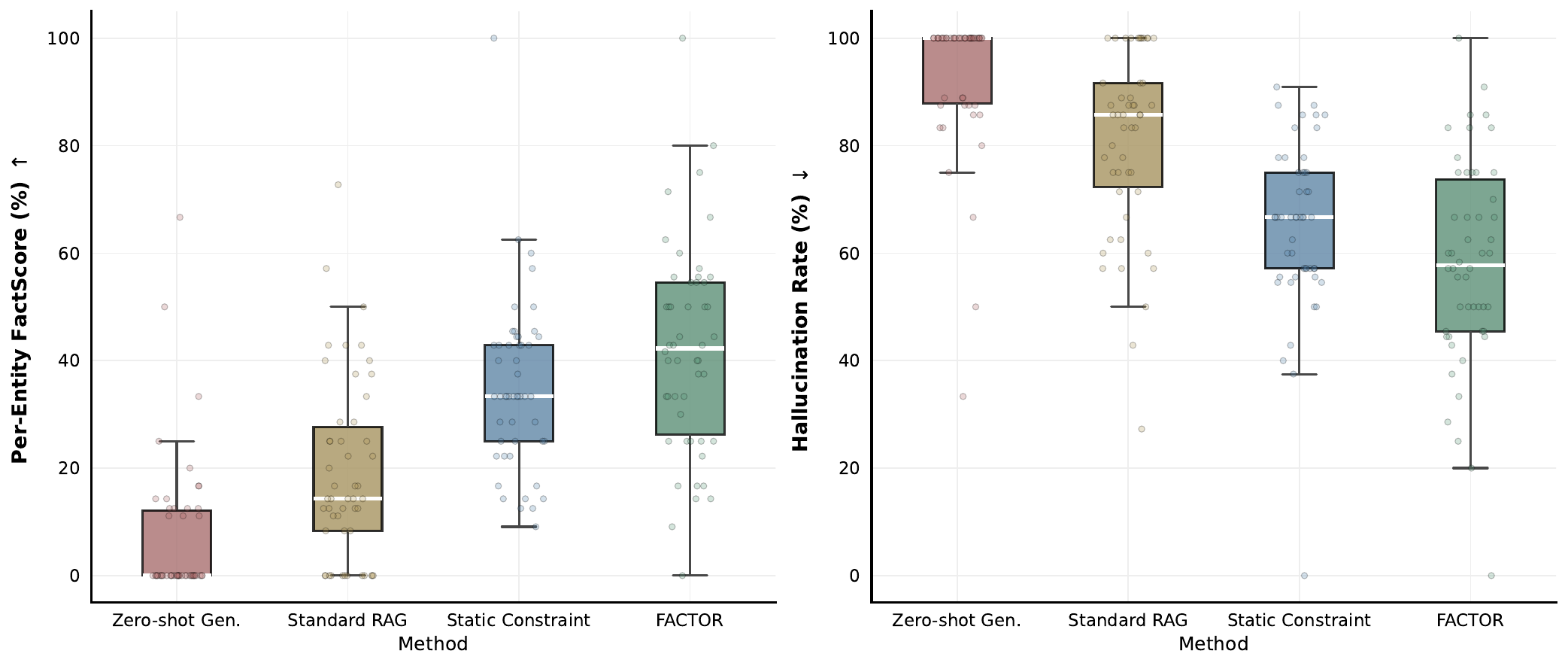}
    \caption{Per-entity FactScore and hallucination-rate distributions for all methods. Median factuality and consistency across the different topics in the evaluation improves for FACTOR.}
    \label{fig:boxplot}
\end{figure}

The results of the Wilcoxon signed-rank tests are statistically significant and shows improvements over Zero-shot Generation, Standard RAG, and Static Verification. Figure~\ref{fig:boxplot} also shows that FACTOR achieves more uniform factuality ratings within entities; and reducing hallucination rates.

\begin{table}[ht]
\centering
\scriptsize
\setlength{\tabcolsep}{5pt}
\renewcommand{\arraystretch}{1.15}
\begin{tabular}{@{}lccc@{}}
\toprule
\textbf{Method}
  & \textbf{Avg Ver.\ Calls $\downarrow$}
  & \textbf{Calls/Claim $\downarrow$}
  & \textbf{Gen.\ Time (s) $\downarrow$} \\
\midrule
Zero-shot Gen. & 44.7  & 4.92  & 12.5 \\
Standard RAG   & 52.4  & 4.92  & 15.6 \\
Static Constr. & 194.0 & 20.95 & 22.8 \\
FACTOR   & \textbf{41.9}  & \textbf{4.92}  & 51.0 \\
\bottomrule
\end{tabular}
\caption{Generation latency and verification cost for all methods. FACTOR does not decrease verification effort as much as the static verification, but it does do so at higher inference latency.}
\label{tab:5}
\end{table}

\subsection{Efficiency Analysis}
The verification cost and the generation latency are reported in Table~\ref{tab:5}. In comparison to Static Verification, FACTOR significantly decreases the number of verification calls, and improves factuality. This reduction is due to uncertainty-aware routing, a lightweight verification for low-risk content and strict verification for uncertain content.

\begin{figure}[ht]
    \centering
    \includegraphics[width=\columnwidth]{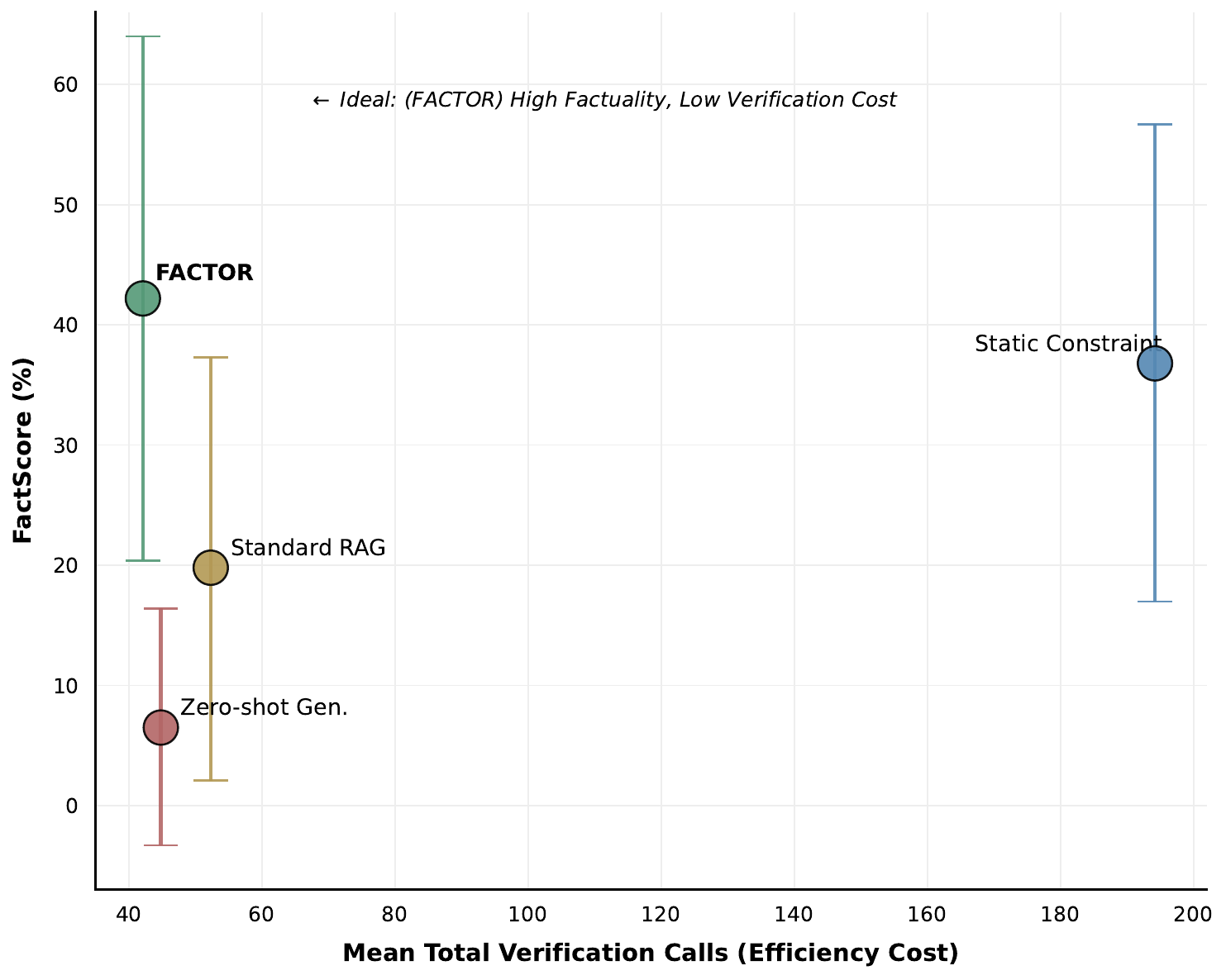}
    \caption{Efficiency--factuality trade-off across methods. Factuality is increased while the verification cost is significantly reduced with FACTOR compared to static verification.}
    \label{fig:pareto}
\end{figure}

Efficient does not directly mean "fast". FACTOR adds new overhead associated with uncertainty estimation and sampling of candidates, increasing generation times. This trade-off is okay in situations where accuracy of information is more important than the interactive speed of response. This balance is illustrated by Figure~\ref{fig:pareto}: FACTOR is in a more attractive factuality--efficiency region than the other methods.

\begin{table}[ht]
\centering
\scriptsize
\setlength{\tabcolsep}{5pt}
\renewcommand{\arraystretch}{1.15}
\begin{tabular}{@{}lcc@{}}
\toprule
\textbf{Method}
  & \textbf{Mean PPL $\downarrow$}
  & \textbf{FactScore $\uparrow$} \\
\midrule
Zero-shot Gen. & 1.8 &  6.5 \\
Standard RAG   & 6.1 & 19.7 \\
Static Constr. & 5.5 & 36.8 \\
FACTOR         & 5.5 & \textbf{42.3} \\
\bottomrule
\end{tabular}
\caption{Fluency and factuality in all methods. Compared with the static verification, FACTOR has the ability to enhance the factuality without reducing the output fluency.}
\label{tab:6}
\end{table}

\subsection{Fluency Analysis}
Table~\ref{tab:6} lists the perplexity as an approximation of fluency. Zero-shot Generation has the lowest perplexity because it does not impose any constraints on decoding. Its factuality, however, is still significantly less than retrieval- and verification-based methods, suggesting that fluent text is not always factual.

Despite the differences in factuality, FACTOR and Static Verification have similar perplexities. This indicates that it is possible to adaptively verify factual consistency without compromising output quality. The fluency item in the re-ranking objective thus maintains readability while factuality is the basis for candidate selection.


\begin{table}[ht]
\centering
\scriptsize
\setlength{\tabcolsep}{5pt}
\renewcommand{\arraystretch}{1.15}
\begin{tabular}{@{}lcc@{}}
\toprule
\textbf{Configuration}
  & \textbf{FactScore $\uparrow$}
  & \textbf{Avg Ver.\ Calls $\downarrow$} \\
\midrule
Entropy only                & 42.9 & 39.2 \\
Consistency only     & \textbf{50.9} & \textbf{36.7} \\
Combined equal              & 46.4 & 37.1 \\
Entropy-heavy (0.7/0.3)             & 46.3 & 38.4 \\
Consistency-heavy (0.3/0.7)          & 41.0 & 36.9 \\
\bottomrule
\end{tabular}
\caption{The impact of uncertainty-signal weighting on factuality and verification cost. Combined weighting makes for a balanced contribution, whereas semantic consistency makes for a strong contribution to factuality.}
\label{tab:7}
\end{table}

\section{Analysis}\label{sec:ablation}
We perform three ablation studies to separate the contribution of uncertainty estimation, adaptive verification thresholds and claim segmentation.

\subsection{Uncertainty Signals}
Table~\ref{tab:7} compares the various uncertainty-weighting configurations. Semantic consistency is the highest FactScore and shows strong correlation with factual reliability when compared to local level of confidence.Nevertheless, the equal weight configuration is a better compromise between the factuality and the verification cost, so it is used for the main experiments.

\begin{figure}[h]
    \centering
    \includegraphics[width=\columnwidth]{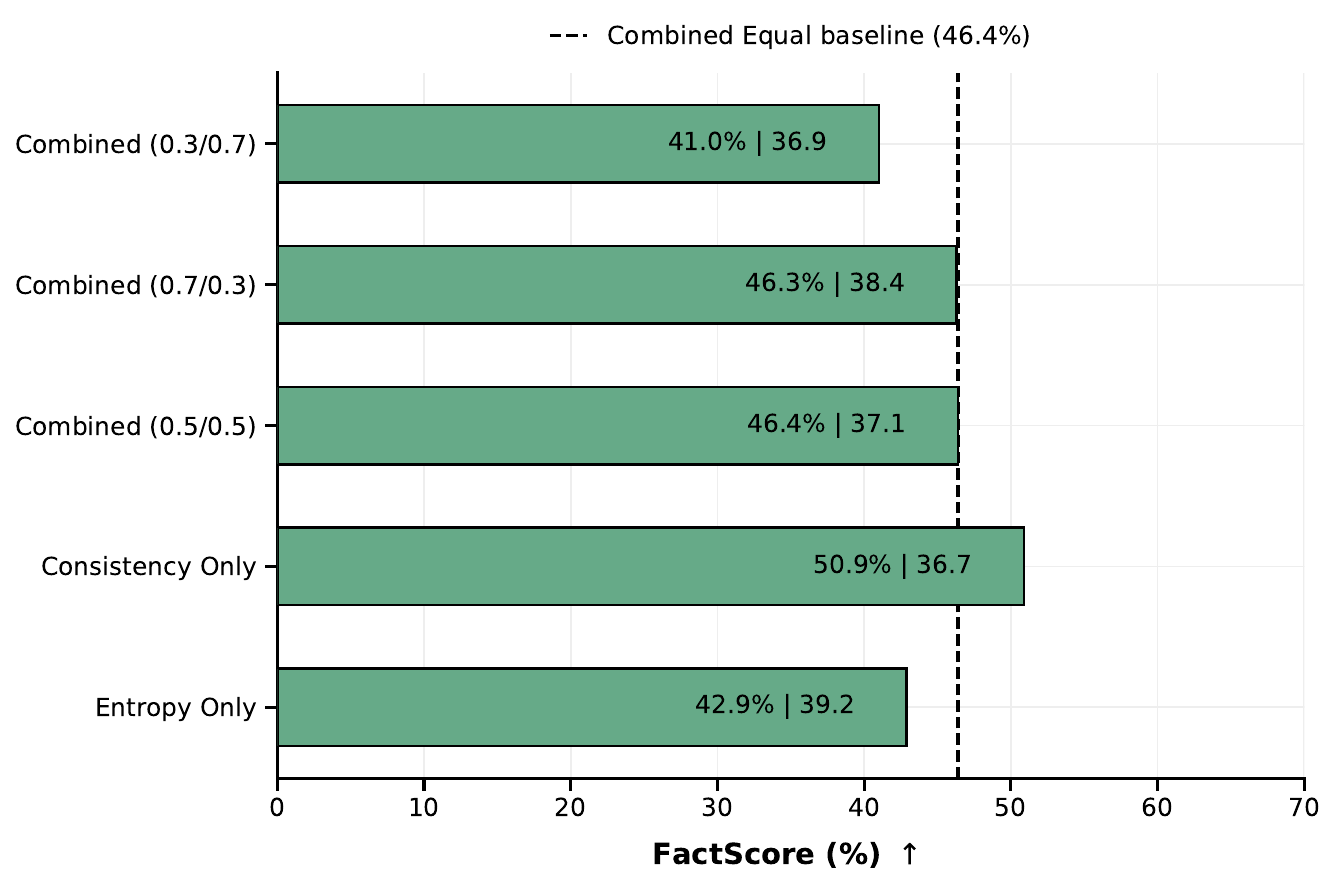}
    \caption{Uncertainty-weighting configurations with 95\% confidence intervals across FactScore.}
    \label{fig:uncertainty_components}
\end{figure}

This trend is further reflected in Figure~\ref{fig:uncertainty_components} which looks at the performance differences between topics. Consistency-only routing results in the higher average FactScore, but with relatively large confidence intervals. The equal weight configuration, however, provides comparable results without compromising much in terms of factuality. In total, these results suggest that entropy and semantic consistency provide complementary signals of uncertainty, where entropy makes the routing process more robust, and consistency leads to gains in factuality, to a greater degree.

\begin{table}[ht]
\centering
\scriptsize
\setlength{\tabcolsep}{5pt}
\renewcommand{\arraystretch}{1.15}
\begin{tabular}{@{}lccc@{}}
\toprule
\textbf{Configuration}
  & \textbf{Low / High}
  & \textbf{FactScore  $\uparrow$}
  & \textbf{Ver.\ Calls $\downarrow$} \\
\midrule
Very tight & 0.10 / 0.40 & 36.9 & 37.5 \\
Tight      & 0.20 / 0.50 & 41.4 & 37.0 \\
Default     & 0.30 / 0.70 & 40.9          & \textbf{37.1} \\
Relaxed     & 0.40 / 0.80 & \textbf{47.9} & 38.6 \\
\bottomrule
\end{tabular}
\caption{Uncertainty-tier boundaries and factuality and verification cost. The performance of the verifiers is strongly influenced by the way the uncertainty is expressed as verification criteria.}
\label{tab:8}
\end{table}
\subsection{Threshold Sensitivity}
Different uncertainty-tier boundaries were evaluated in Table~\ref{tab:8}. Loose thresholds get the best FactScore, while very tight thresholds significantly decrease FactScore. If most claims are directed to more rigorous verification levels, the system becomes more like uniform verification and does not reap the gains of adaptive routing.

\begin{figure}[h]
    \centering
    \includegraphics[width=\columnwidth]{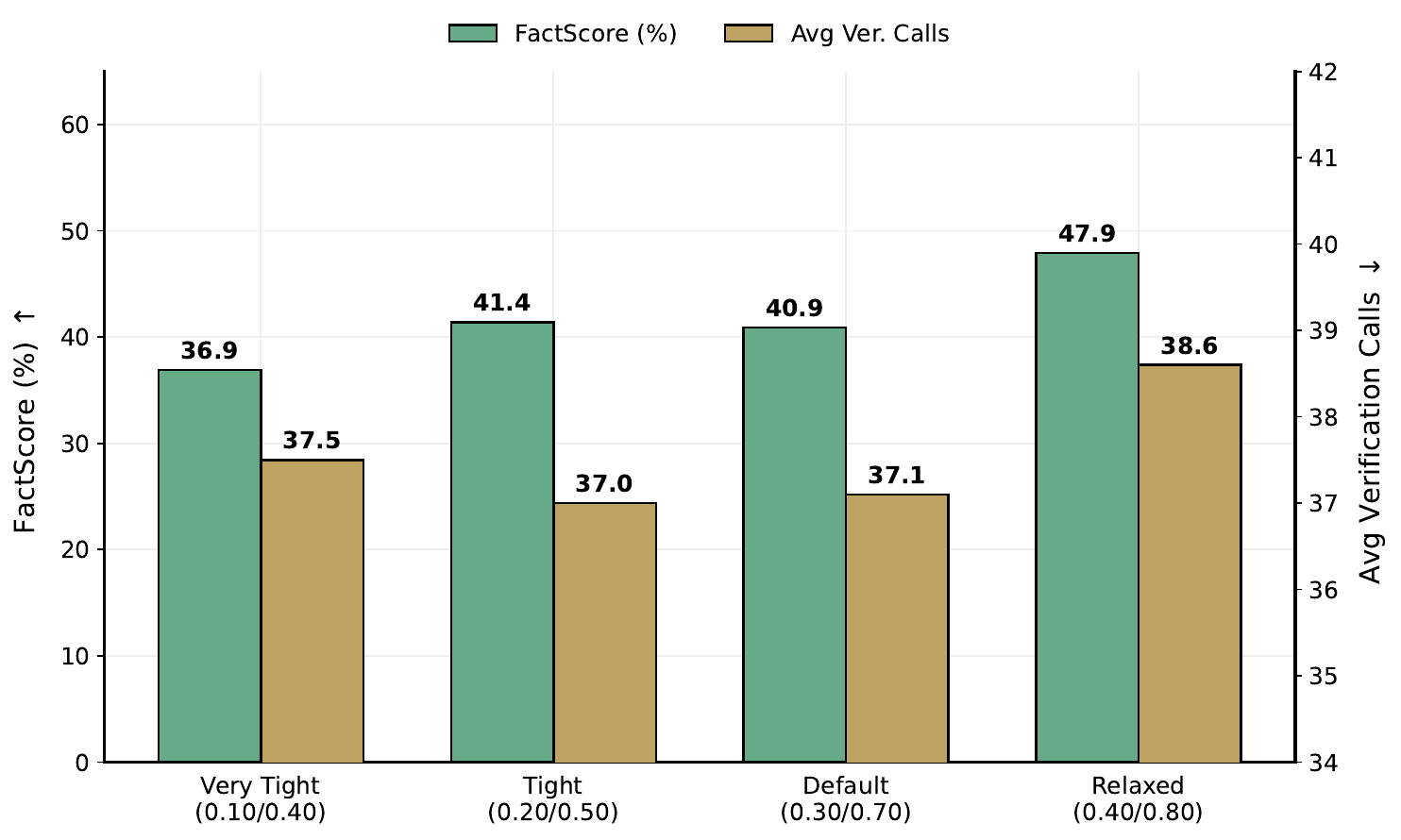}
    \caption{The uncertainty-tier configurations with the factScore and verification cost.}
    \label{fig:threshold_sensitivity}
\end{figure}

The relationship between factuality and verification cost are shown in Fig.~\ref{fig:threshold_sensitivity}. The default thresholds appear to be somewhat conservative, with the relaxed configuration providing better factuality with little additional verification effort. More globally, the results validate the fact that the effectiveness of verification is linked to the way uncertainty is translated in the verification criteria.

\begin{table}[ht]
\centering
\scriptsize
\setlength{\tabcolsep}{5pt}
\renewcommand{\arraystretch}{1.15}
\begin{tabular}{@{}lccc@{}}
\toprule
\textbf{Granularity}
  & \textbf{FactScore $\uparrow$}
  & \textbf{Avg Claims $\downarrow$}
  & \textbf{Avg Ver.\ Calls $\downarrow$} \\
\midrule
Sentence & \textbf{46.3} & \textbf{6.93} & \textbf{34.4} \\
Atomic   & 41.5 & 7.60 & 35.9 \\
\bottomrule
\end{tabular}
\caption{The impact of segmenting claims in detail. Sentence-level claims are more factually strong and less costly to verify than atomic claims decomposition.}
\label{tab:9}
\end{table}

\subsection{Claim Granularity}
In Table~\ref{tab:9} a comparison of sentence-level and atomic claim segmentation is provided. Claims at sentence level have a higher fact value and fewer callouts for verification. Atomic segmentation is useful to improve granularity but can lead to loss of contextual information required for reliable entailment assessment.

\begin{figure}[h]
    \centering
    \includegraphics[width=\columnwidth]{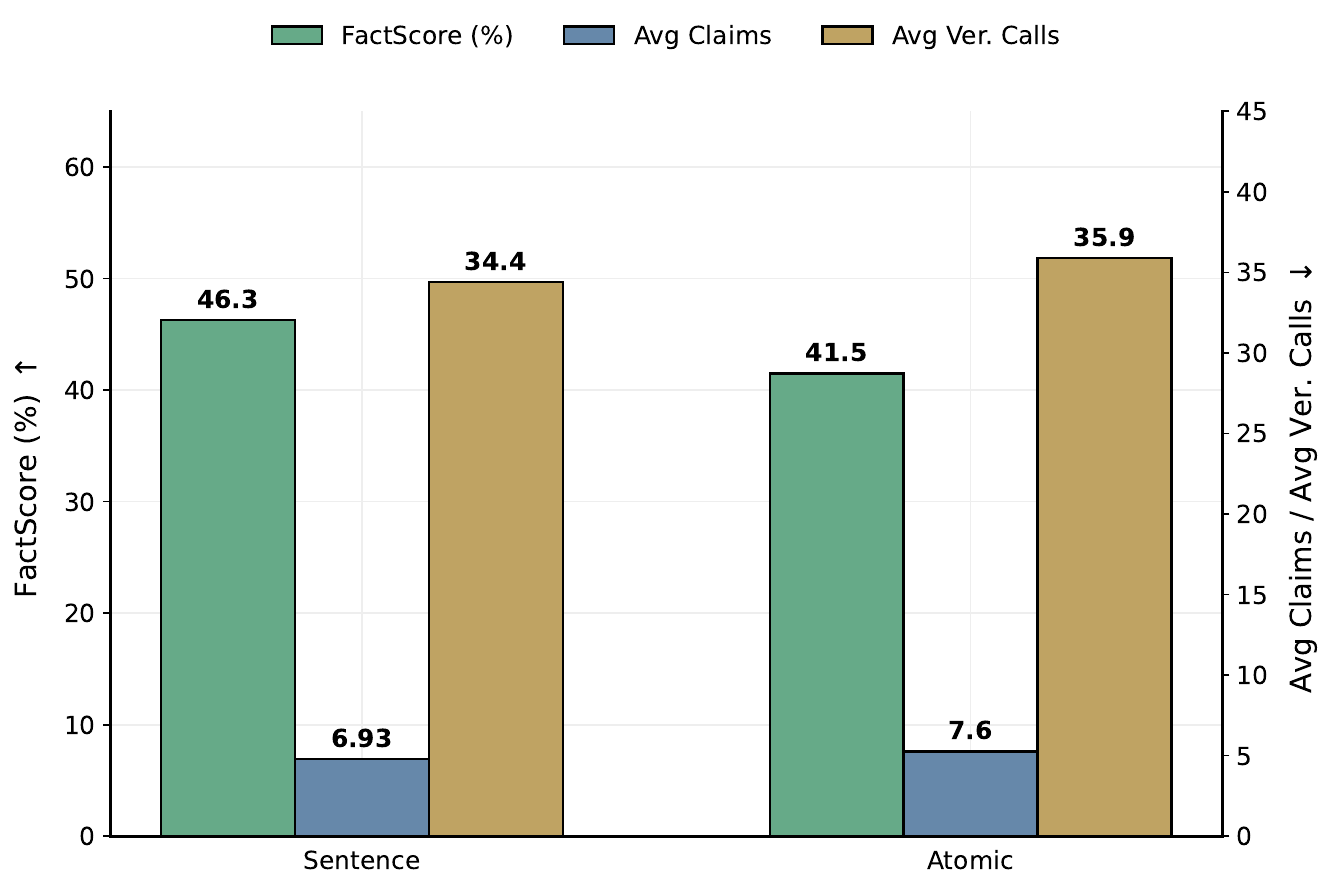}
    \caption{FactScore and normalized verification cost for sentence-level and atomic claim segmentation.}
    \label{fig:claim_granularity}
\end{figure}

The trade-off between factuality and efficiency of the two segmentation strategies is illustrated in Figure~\ref{fig:claim_granularity}. Sentence-level verification consistently dominates atomic decomposition, suggesting that preserving contextual information is more beneficial than increasing claim granularity. These results support the use of sentence-level claims throughout the remainder of the study.

\section{Conclusions}
The paper investigates that how claims within a generated response are at different hallucination risk levels, yet existing verification techniques mostly treat them the same. We address this problem with FACTOR that adapts verification criteria according to claim-level uncertainty. Our empirical investigation shows that uncertainty-aware verification improves factuality while reducing verification cost relative to static verification strategies in the response generation. These findings suggest that factuality depends not only on retrieval quality or verification strength, but also on how verification effort is allocated. Treating verification as a risk-aware process offers a simple and model-agnostic path toward more reliable long-form generation.
 
\bibliographystyle{IEEEtran}
\bibliography{references}
 
\end{document}